\documentclass[10pt,twocolumn,letterpaper]{article}

\usepackage{iccv}
\usepackage{times}
\usepackage{epsfig}
\usepackage{graphicx}
\usepackage{amsmath}
\usepackage{amssymb}
\usepackage{url}

\usepackage[pagebackref=true,breaklinks=true,letterpaper=true,colorlinks,bookmarks=false]{hyperref}

\iccvfinalcopy

\newcommand{\ourmethodfull}[0]{Logit Regularization Methods}
\newcommand{\ourmethod}[0]{LRM}

\def\iccvPaperID{4388}

\ificcvfinal\fi

\begin{document}

\title{Improved Adversarial Robustness via Logit Regularization Methods}

\author{
  {\normalfont Cecilia Summers} \\
  Department of Computer Science\\
  University of Auckland\\
  {\tt\small cecilia.summers.07@gmail.com} \\
\and
  Michael J. Dinneen \\
  Department of Computer Science\\
  University of Auckland\\
  {\tt\small mjd@cs.auckland.ac.nz} \\
}

\maketitle

\begin{abstract}
While great progress has been made at making neural networks effective across a wide range of visual tasks, most models are surprisingly vulnerable.
This frailness takes the form of small, carefully chosen perturbations of their input, known as adversarial examples, which represent a security threat for learned vision models in the wild -- a threat which should be responsibly defended against in safety-critical applications of computer vision. In this paper, we advocate for and experimentally investigate the use of a family of logit regularization techniques as an adversarial defense, which can be used in conjunction with other methods for creating adversarial robustness at little to no marginal cost. We also demonstrate that much of the effectiveness of one recent adversarial defense mechanism can in fact be attributed to logit regularization, and show how to improve its defense against both white-box and black-box attacks, in the process creating a stronger black-box attack against PGD-based models. We validate our methods on three datasets and include results on both gradient-free attacks and strong gradient-based iterative attacks with as many as 1,000 steps.
\end{abstract}

\section{Introduction}

Neural networks, despite their high performance on a variety of tasks, can be brittle. Given data intentionally chosen to trick them, many deep learning models suffer extremely low performance. This type of data, commonly referred to as \emph{adversarial examples}, represent a security threat to any machine learning system where an attacker has the ability to choose data input to a model, potentially allowing the attacker to control a model's behavior.

At the same time, applications of computer vision are pervasive, with future applications including autonomous driving and medical diagnostics.
While these use cases of vision are exciting in their potential for societal good, they also have the potential to be grave threats when behaving erroneously, with undesired behavior able to cause harm to both their users and creators.
It is therefore of critical importance to understand how to defend against such adversarial attacks, both to prevent these systems from failing and to prevent malicious actors from exploiting any vulnerabilities they may have.
Though challenging, this is nonetheless an urgent need, as it is already known that attacks targeting systems for autonomous driving and medical diagnostics are possible~\cite{eykholt2018robust,Finlayson1287}.

Today, adversarial examples are typically created by small, carefully chosen transformations of data that models are otherwise high-performant on.
While this is primarily due to the ease of experimentation with existing datasets~\cite{gilmer2018motivating}, the full threat of adversarial examples is indeed only limited by the ability and creativity of an attacker's example generation process -- for example, even relatively basic research has shown the potential for adversarial attacks in the physical world~\cite{kurakin2016adversarial}, with more attacks being found on a regular basis.

Even with the limited threat models considered in current research, performance on adversarially chosen examples can be dramatically worse than unperturbed data. One canonical example is the CIFAR-10 image classification task~\cite{krizhevsky2009learning}, where white-box accuracy on adversarially chosen examples is lower than $50\%$, even for the most robust defenses known today~\cite{madry2018towards,kannan2018adversarial}, while unperturbed accuracy can be as high as $98.5\%$~\cite{cubuk2018autoaugment}, a $30\times$ difference in misclassification rate. On larger tasks, such as ImageNet~\cite{russakovsky2015imagenet}, the difference is even bigger, as no model is known to be robust to any but the weakest of all adversarial attacks.

Current defenses against adversarial examples generally come in one of a few flavors. Perhaps the most common approach is to generate adversarial examples as part of the training procedure and explicitly train on them, known as ``adversarial training''. Another approach is to transform the model's input representation in a way that thwarts an attacker's adversarial example construction mechanism. 
While these methods can be effective, care must be taken to make sure that they are not merely obfuscating gradients~\cite{athalye2018obfuscated}.
Last, generative models can be built to model the original data distribution, recognizing when the input data is out of sample and potentially correcting it~\cite{song2018pixeldefend,samangouei2018defense}.
Of these, arguably the most robust defenses today follow the adversarial training paradigm, of which adversarial logit pairing~\cite{kannan2018adversarial} is the most recent incarnation, extending the adversarial training work of Madry \emph{et al.}~\cite{madry2018towards} by incorporating an additional term to make the logits (pre-softmax values) of an unperturbed and adversarial example more similar. 

In this work, we show that adversarial logit pairing derives a large fraction of its benefits from regularizing the model's logits toward zero, which we demonstrate through simple and easy to understand theoretical arguments in addition to empirical demonstration. Investigating this phenomenon further, we examine two alternatives for logit regularization, finding that both result in improved robustness to adversarial examples, sometimes surprisingly so -- for example, using the right amount of label smoothing~\cite{szegedy2016rethinking} can result in greater than $40\%$ robustness to a 10-step projected gradient descent (PGD) attack~\cite{madry2018towards} on CIFAR-10 while training only on the original, unperturbed training examples, and is also a compelling black-box defense.  We then present an alternative formulation of adversarial logit pairing that separates the logit pairing and logit regularization effects, improving the defense. The end result of these investigations is a defense that outperforms state-of-the-art approaches for PGD-based adversaries on CIFAR-10 for both white-box and black-box attacks, while requiring little to no computational overhead on top of adversarial training.

\section{Overview of Adversarial Training}
Before proceeding with our analysis, we review existing work on adversarial training for context. While adversarial examples have been examined in the machine learning community in some capacity for many years~\cite{dalvi2004adversarial}, their study has drawn a sharp focus in the current renaissance of deep learning, starting with Szegedy \emph{et al.}~\cite{szegedy2014intriguing} and Goodfellow \emph{et al.}~\cite{goodfellow2015explaining}, particularly in the context of computer vision. In Goodfellow \emph{et al.}~\cite{goodfellow2015explaining}, adversarial training is presented as training with a weighted loss between an original and adversarial example, \emph{i.e.} with a loss of

\begin{equation}
\begin{split}
  \label{eq:adversarial_training}
  \tilde{J}(\theta, x, y) = \frac{1}{m} \sum_{i=1}^m & \alpha J(\theta, x^{(i)}, y^{(i)}) + \\ & (1 - \alpha) J(\theta, g(x^{(i)}), y^{(i)})
\end{split}
\end{equation}
where $g(x)$ is a function representing the adversarial example generation process, originally presented as $g(x) = x + \epsilon \cdot \text{sign}(\nabla_x J(\theta, x, y))$, $\alpha$ is a weighting term between the original and adversarial examples typically set to $0.5$, $\theta$ are the model parameters to learn, $J$ is a cross-entropy loss, $m$ is the dataset size, $x^{(i)}$ is the $i$th input example, and $y^{(i)}$ is its label. Due to the use of a single signed gradient with respect to the input example, this method was termed the ``fast gradient sign method'' (FGSM), requiring a single additional forward and backward pass of the network to create. Kurakin \emph{et al.}~\cite{kurakin2016adversarial} extended FGSM into a multi-step attack, iteratively adjusting the perturbation applied to the input example through several rounds of FGSM. This was also the first attack that could be described as a variant of projected gradient descent (PGD), where the adversarial perturbation is initialized to zero. Both of these approaches primarily target an $L^\infty$ threat model, where the $L^\infty$ norm between the original and adversarial example is constrained to a small value. By keeping the $L^\infty$ norm small, it is assumed that the adversarial example will have the same correct label as the original example, \emph{i.e.} that the perturbation is small enough to still be easily recognizable as the original category, an assumption that allows for research in the field without requiring the manual annotation of every new adversarial example.

Madry \emph{et al.}~\cite{madry2018towards} built upon these works by initializing the search process for the adversarial perturbation randomly, and is among the strongest attacks currently available.
Although only a slight modification of Kurakin \emph{et al.}~\cite{kurakin2016adversarial}, this detail is critical -- with a zero initialization it is easy to become robust only at existing training points, thus causing a ``gradient masking'' effect~\cite{athalye2018obfuscated}.
Through extensive experiments, they showed that even performing PGD with a single random initialization is able to approximate the strongest adversary found with current first-order methods, and doing adversarial training with this attack resulted in the most robust model yet.
However, as with multi-step FGSM, performing adversarial training with this approach can be rather expensive, taking an order of magnitude longer than standard training. Specifically, PGD-based adversarial training requires $N+1$ forward and backward passes of the model, where $N$ is the number of PGD iterations, and is typically on the order of 5 to 20~\cite{madry2018towards}.

Improving on PGD-based adversarial training, Kannan \emph{et al.}~\cite{kannan2018adversarial} introduced adversarial logit pairing (ALP), which adds a term to the adversarial training loss function that encourages the model to have similar logits for original and adversarial examples:

\begin{equation}
\begin{split}
  \tilde{J}(\theta, x, y) = \frac{1}{m} \sum_{i=1}^m & \alpha J(\theta, x^{(i)}, y^{(i)}) + \\ & (1 - \alpha) J(\theta, g(x^{(i)}), y^{(i)}) + \\ & \lambda L(f(x^{(i)}; \theta), f(g(x^{(i))}; \theta)).
\end{split}
\end{equation}

where $L$ was set to an $L^2$ loss and $f(x, \theta)$ returns the logits of the model corresponding to example $x$. Adversarial logit pairing has the motivation of increasing the amount of structure given to the model in the learning process by encouraging the model to have similar prediction patterns on the original and adversarial examples, a process reminiscent of distillation~\cite{hinton2015distilling}.

Kannan \emph{et al.}~\cite{kannan2018adversarial} also studied a baseline version of ALP, called ``clean logit pairing'', which paired randomly chosen unperturbed examples together. Surprisingly, this worked reasonably well, inspiring them to experiment with a similar idea they call ``clean logit squeezing'', regularizing the $L^2$ norm of the model's logits, which worked even more effectively, though this idea itself was not combined with adversarial training. It is this aspect of the work that is most related to what we study in this paper.

Last, it is worth noting work examining the reproducibility of adversarial logit pairing~\cite{engstrom2018evaluating}. While it is true that ALP was found to not actually be robust on ImageNet to multi-step white-box attacks with a large number of iterations, continuing the trend of no models being robust on ImageNet, the improved robustness of ALP on smaller datasets that are more commonly used was not refuted, a fact which we also find, even with attacks of up to 1,000 steps. Thus, we believe that ALP shows some promise in advancing our understanding and effectiveness in defending against adversarial examples.

\section{Adversarial Logit Pairing and Logit Regularization}

We now show how adversarial logit pairing~\cite{kannan2018adversarial} acts as a logit regularizer.
For notational convenience, denote $\ell_c^{(i)}$ as the logit of the model for class $c$ on example $i$ in its original, unperturbed form, and $\tilde{\ell}_c^{(i)}$ as the logit for the corresponding adversarial example. The logit pairing term in adversarial logit pairing is a simple $L^2$ loss:

\begin{equation}
  L = \frac{1}{2} (\ell^{(i)}_c - \tilde{\ell}^{(i)}_c)^2 \\
\end{equation}

While it is obvious that minimizing this term will have the effect of making the original and adversarial logits more similar in some capacity, what precise effect does it have on the model during training?
To examine the effect of such a loss in gradient-based training, the dominant training paradigm for almost all computer vision models today, we can look at the gradient of this loss term with respect to the logits themselves:

\begin{equation}
  \frac{\partial L}{\partial \ell^{(i)}_c} = \ell^{(i)}_c - \tilde{\ell}^{(i)}_c
\end{equation}

\begin{equation}
  \frac{\partial L}{\partial \tilde{\ell}^{(i)}_c} =  \tilde{\ell}^{(i)}_c - \ell^{(i)}_c
\end{equation}

Under the assumption that the adversarial example moves the model's predictions away from the correct label (as should be the case with any reasonable adversarial example, such as an untargeted PGD-based attack), we will have that $\ell^{(i)}_c > \tilde{\ell}^{(i)}_c$ when $c = y^{(i)}$ is the correct category, and $\ell^{(i)}_c < \tilde{\ell}^{(i)}_c$ otherwise\footnote{While it is theoretically possible that the logits for an incorrect category are smaller in a loss-maximizing adversarial example, increasing their value, all else equal, will result in a higher loss, so $\ell^{(i)}_c < \tilde{\ell}^{(i)}_c$ will typically hold.}. Keeping in mind that model updates move in the direction opposite of the gradient, then the update to the model's weights will attempt to make the original logits smaller and the adversarial logits larger when $c = y^{(i)}$ and will otherwise attempt to make the original logits larger and the adversarial logits smaller.

However, it is not sufficient to examine this in isolation, as logit pairing is only one component of a typical adverarial loss: it must be considered in the context of the adversarial training loss $\tilde{J}$ -- in particular, the cross-entropy loss used in $\tilde{J}$ for the adversarial example $g(x^{(i)})$ already encourages the adversarial logits to be higher for the correct category and smaller for all incorrect categories, and furthermore the scale of the loss $\tilde{J}$ typically is an order of magnitude larger than the adversarial pairing loss. Thus, we argue that the main effect of adversarial logit pairing is actually in the remaining two types of updates, encouraging the logits of the original example to be smaller for the correct category and larger for all incorrect categories -- an effect which is essentially regularizing model logits in a manner similar to ``logit squeezing''~\cite{kannan2018adversarial} or label smoothing~\cite{szegedy2016rethinking}.

Examining this further, we can also take a different perspective by explicitly incorporating the \emph{scale} of the logits in the logit pairing term.
If we factor out a shared scale factor $\gamma$ from each logit, the logit pairing term has the form

\begin{equation}
  L = \frac{1}{2} (\gamma \ell^{(i)}_c - \gamma \tilde{\ell}^{(i)}_c)^2 \\
\end{equation}
implying that
\begin{equation}
  \frac{\partial L}{\partial \gamma} = \gamma (\ell^{(i)}_c - \tilde{\ell}^{(i)}_c)^2,  \\
\end{equation}
Since $(\ell^{(i)}_c - \tilde{\ell}^{(i)}_c)^2$ is non-negative, this means that the model will always attempt to update the scale $\gamma$ of the logits in the opposite direction of its sign, which is necessarily an update moving $\gamma$ toward zero so long as the logits were not identical -- in fact, if this were the only term in the loss, then it is easy to see that $\gamma = 0$ would be a global minimizer of the loss.
However, in practice this affect is partially counterbalanced by the adversarial training term, which requires that logits across different categories be different in order to minimize its cross-entropy loss.

Given this interpretation, in this work we now explore four key questions: 1) Experimental verification of our analysis. In practice, how much of a logit regularization effect does ALP have?
2) Do other forms of logit regularization have similar effects on adversarial robustness? If so, then an entire family of methods for improving adversarial robustness will have been found.
3) Is it possible to decouple adversarial logit pairing explicitly into a form where the effect of logit regularization and pairing can be disentangled?
4) Finally, using these insights, can we discover even more robust models?

\subsection{Experimental Evidence}
\begin{figure}[t]
\begin{center}
  \includegraphics[width=.49\linewidth]{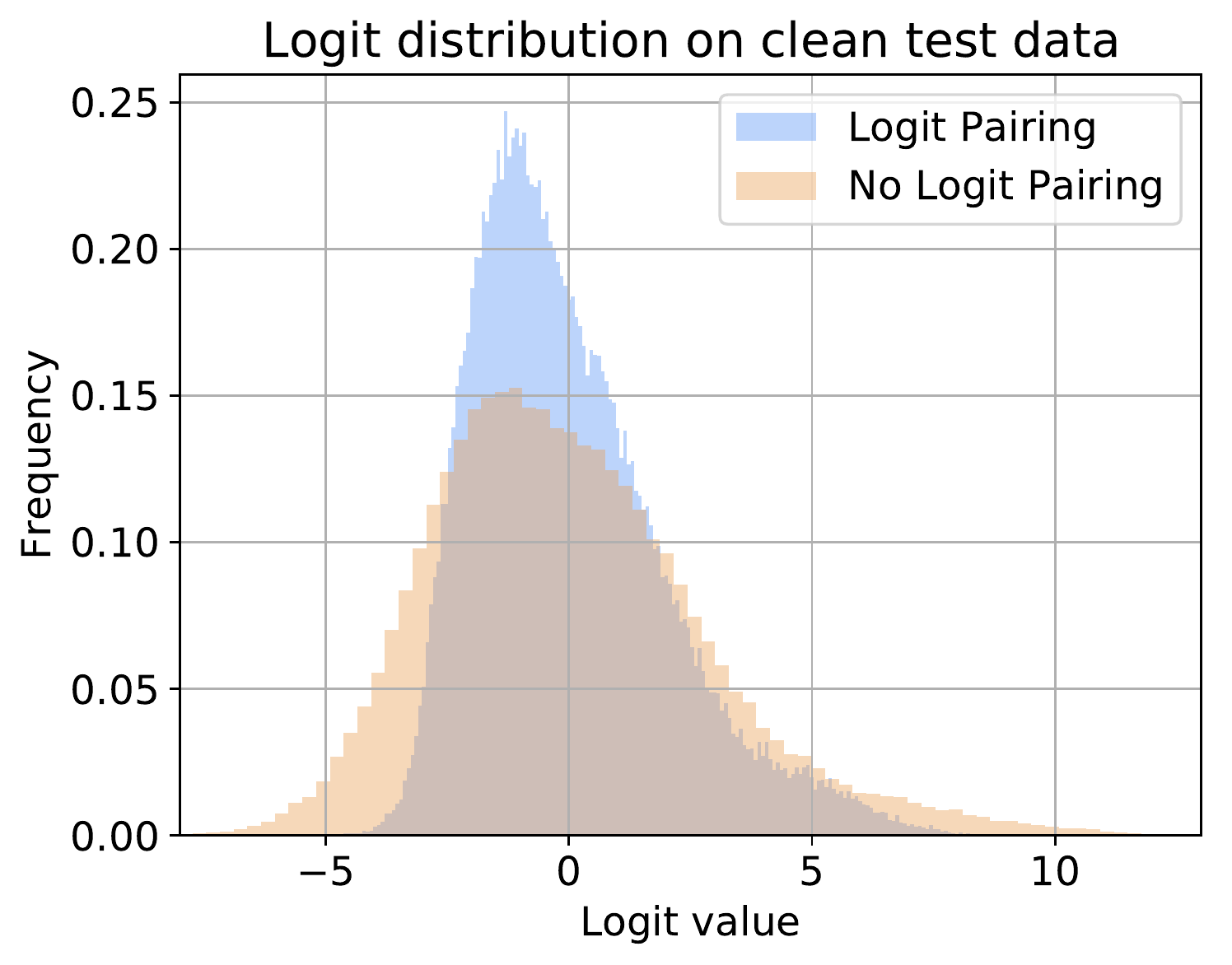}
  \includegraphics[width=.49\linewidth]{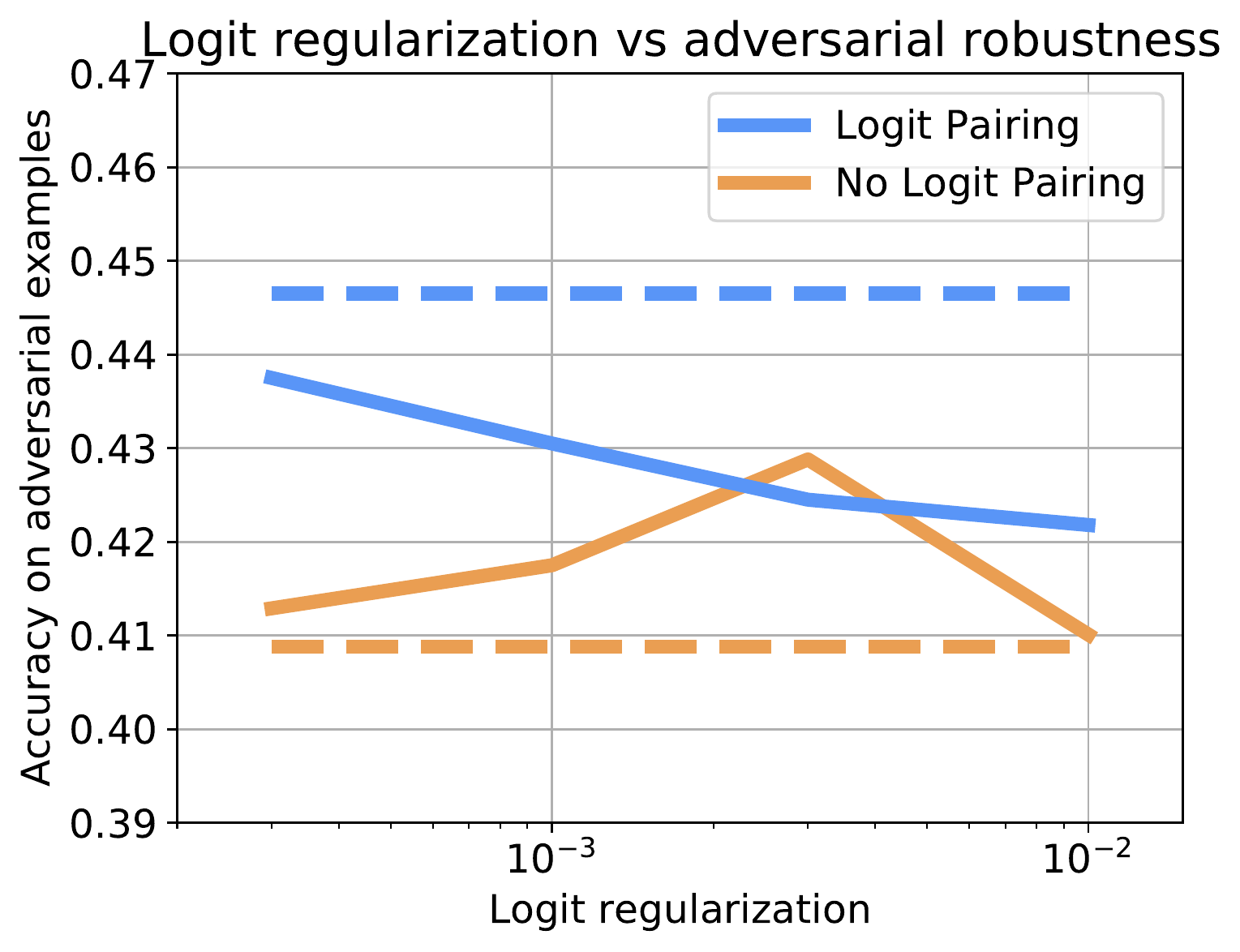}
\end{center}
\caption{
  Left: Distribution of logits on clean test data for models trained with and without logit pairing.
  Right: Performance against a 10-step PGD attack for models trained with varying amounts of logit regularization, with and without logit pairing.
}
  \label{fig:alp_vs_pgd}
\end{figure}

Perhaps the most straightforward way to test our hypothesis is to examine the logits of a model trained with ALP vs one trained with standard adversarial training. If our hypothesis is true, then the model trained with ALP will have logits that are generally smaller in magnitude.
We present the results of this experiment in Figure~\ref{fig:alp_vs_pgd}(left), using an 18-layer ResNet~\cite{he2016deep} classifier trained on CIFAR-10~\cite{krizhevsky2009learning} as our experimental testbed (see Sec.~\ref{sec:implementation_details} for details).

As shown in Figure~\ref{fig:alp_vs_pgd}, it is indeed the case that the logits for a model trained with ALP are of smaller magnitude than those of a model trained with PGD, with a variance reduction of the logits from $8.31$ to $4.02$ on clean test data\footnote{The corresponding distributions on a set of PGD-based adversarial examples are very similar.}.
Though this provides evidence that ALP \emph{does} have the effect of regularizing logits, this data alone is not sufficient to determine if logit regularization is a key mechanism in ALP's improved adversarial robustness.

To answer this, we examine if standard adversarial training can be improved by explicitly regularizing the logits. If adversarial robustness can be improved, but similar improvements can \emph{not} be made to ALP, then at least some of the benefits of ALP can be attributed strictly to logit regularization.
We present the results of this experiments in Figure~\ref{fig:alp_vs_pgd}(right), implemented using the ``logit squeezing'' form of regularization ($L^2$-regularization on the logits).

As shown, we find that incorporating regularization on model logits is able to recover slightly more than half of the total improvement from logit pairing, with a unimodal distribution -- too little regularization has only a small effect, and too much regularization approaches the point of being harmful.
In contrast, when added to a model already trained with ALP, regularizing the logits does not lead to any improvement at all, and in fact hurts performance at all levels of regularization strength, likely due to the combination of explicit logit regularization and the implicit regularization happening from ALP overpowering the cross-entropy loss of adversarial training.
This evidence makes clear that one of the key improvements from logit pairing is due to a logit regularization effect.

We would like to emphasize that these results are not meant to diminish ALP in any sense -- our goals are to investigate the mechanism by which it works and explore if it can be generalized or improved.
Thus, given these results, it is worth examining other methods that have an effect of regularizing logits in order to tell whether this is a more general phenomenon.

\section{Other forms of logit regularization}

\paragraph{Label Smoothing.}
Label smoothing is the process of replacing the one-hot training distribution of labels with a softer distribution, where the probability of the correct class has been smoothed out onto the incorrect classes~\cite{szegedy2016rethinking}. Concretely, label smoothing uses the target distribution:
\begin{equation}
    p^{(i)}_c = 
  \begin{cases}
    1 - s & c = y^{(i)} \\
    \frac{s}{C-1} & c \neq y^{(i)} \\
  \end{cases}
\end{equation}
where $p_c^{(i)}$ is the target probability for class $c$ for example $i$, the number of categories is denoted by $C$, and $s \in [0, 1 - \frac{1}{C}]$ is the smoothing strength.
Label smoothing was originally introduced as a form of regularization, designed to prevent models from being too confident about training examples, and had the goal of improved generalization.
Furthermore, it can be easily implemented as a preprocessing step on the labels, and does not affect model training time in any significant way.
Interestingly, Kurakin \emph{et al.}~\cite{kurakin2016adversarial} found that incorporating a small amount of label smoothing present in a model trained on ImageNet actually \emph{decreased} adversarial robustness roughly by $1\%$. Here we find a different effect.

\begin{figure*}[t]
\begin{center}
  \includegraphics[width=.90\linewidth]{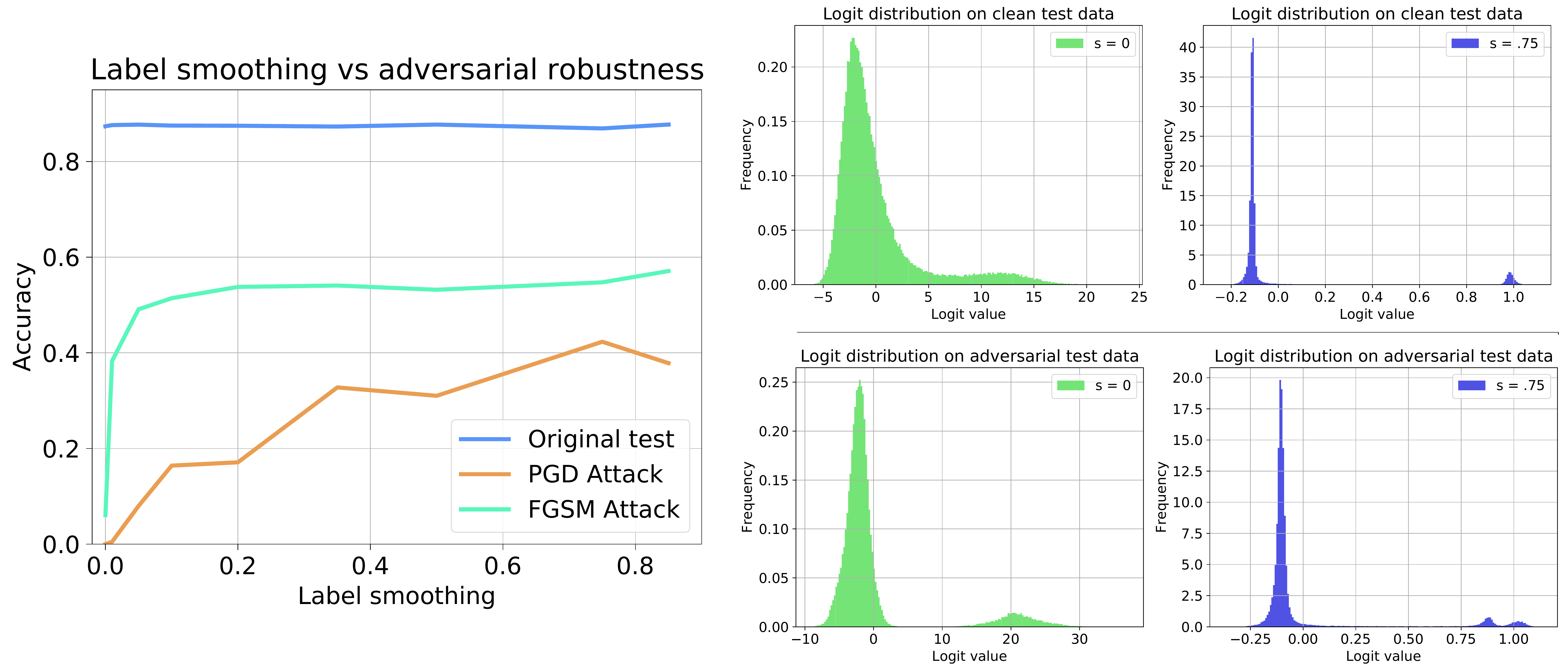}
\end{center}
\caption{
  Left: Clean and adversarial accuracy on CIFAR-10 as a function of the label smoothing strength. Note that models for this figure were trained exclusively on the original training data -- no adversarial examples were involved in the training procedure.
	Top-Right: Logit distribution of model trained with no label smoothing (middle) and with label smoothing of $s = .75$ (right), evaluated on the original test images.
	Bottom-Right: Logit distributions when evaluated on PGD-based adversarial examples.
}
  \label{fig:smooth_effect}
\end{figure*}

In Figure~\ref{fig:smooth_effect}(left) we show the effect label smoothing has on the performance of a model trained only on clean (\emph{i.e.} non-adversarial) training data.
Very surprisingly, using only label smoothing was able to produce a model that is nearly as robust to this 10-step PGD attack as models trained with PGD-based adversarial training or adversarial logit pairing, both of which take an order of magnitude more time to train -- though in faireness we note that when PGD and ALP-based models are trained only on adversarial examples rather than a mixture of clean and adversarial data, their robustness exceeds this performance by around $5\%$.
Furthermore, this benefit of label smoothing comes at no significant loss in accuracy on unperturbed test data, while generally adversarial training tends to trade off original vs adversarial performance.
Another curiosity is that adding in any label smoothing at all dramatically improves robustness to FGSM-based adversaries (adding label smoothing of $s = .01$ brought accuracy up from $6.1\%$ to $38.3\%$), while PGD-based attacks saw much more gradual improvement with label smoothing strength.
While remarkable, this property of label smoothing on the loss surface is eludes understanding, warranting further research.

Examining the logits themselves (Figure~\ref{fig:smooth_effect}, right), we see a striking difference between the models -- the model trained with label smoothing both has a dramatically smaller dynamic range of logits -- roughly 1.2 vs. 20, a 16-fold decrease -- and also presents a much more bimodal logit distribution than the model trained without label smoothing. In other words, it has learned to predict extremely consistent values for logits, a property that may contribute to its adversarial robustness.
Anecdotally, we observed that this behavior held for all positive values of $s$, with a stronger effect the higher $s$ was.

This behavior can be explained: when trained with no label smoothing, the cross-entropy loss used in most models encourages model output to be as close to a one-hot distribution as possible, predicting a probability of 1 for the correct category and 0 for all other categories.
When viewed as logits instead of probabilities, this corresponds to pushing the logits for the correct and incorrect categories as far apart as possible.
However, models trained with label smoothing are instead encouraged to produce a soft distribution, with no probabilities too close to either 0 or 1, corresponding to a bounded target logit difference which gets smaller with increasing $s$.

Additional experiments involving label smoothing are given in Section~\ref{sec:additional_experiments}.

\paragraph{Paired-Example Data Augmentation}
Recently, a new form of data augmentation was found that stands in contrast to standard label-preserving data augmentation.
Paired-example data augmentation consists of combining different training examples together, dramatically altering both the appearance of the training examples and their labels.
Introduced concurrently by multiple groups~\cite{zhang2018mixup,inoue2018data,tokozume2018image}, these types of data augmentation typically have the form of element-wise weighted averaging of two input examples (typically images), with the training label also determined as a weighted average of the original two training labels (represented as one-hot vectors).
Besides making target labels soft (\emph{i.e.} not 1-of-$K$) during training time, these methods also encourage models to behave linearly between examples, which may improve robustness to out of sample data.
Interestingly, Zhang \emph{et al.}~\cite{zhang2018mixup} found that this type of data augmentation improved robustness to FGSM-based attacks on ImageNet~\cite{russakovsky2015imagenet}, but Kannan \emph{et al.}~\cite{kannan2018adversarial} found that the method did not improve robustness against a targeted attack with a stronger PGD-based adversary.

Experimentally, we found evidence agreeing with both conclusions -- when applying \emph{mixup}~\cite{zhang2018mixup}, we found a sizeable increase in robustness to FGSM adversaries, going from $6.1\%$ on CIFAR-10 by training without \emph{mixup} to $30.8\%$ with \emph{mixup}, but did not observe a significant change when evaluated against a PGD-based adversary. While robustness to a PGD adversary with only 5 steps increased by a tiny amount (from $0.0\%$ to $0.5\%$), robustness to a 10-step PGD adversary remained at $0\%$. In our experiments, we use \emph{VH-mixup}, the slightly improved version of \emph{mixup} introduced by Summers and Dinneen~\cite{summers2018improved}.

\subsection{Decoupling Adversarial Logit Pairing}
\label{sec:decouple}
While we have now considered alternate methods by which logits can be regularized, at this point it is still not clear exactly how they might be used with or interact with the logit regularization effect of adversarial logit pairing.
Doing so requires decoupling the logit pairing and logit regularization effects of ALP.

In adversarial logit pairing~\cite{kannan2018adversarial}, the logit pairing term is implemented as an $L^2$ loss:

\begin{equation}
  L(f(x^{(i)}; \theta), f(g(x^{(i)}); \theta)) = \|f(x^{(i)}) - f(g(x^{(i)}))\|^2 ,
\end{equation}
though other losses such as an $L^1$ or Huber loss are also possible.
Expanding this creates a form that makes the pairing and regularization terms evident:

\begin{equation}
\begin{split}
  &L(f(x^{(i)}; \theta), f(g(x^{(i)}); \theta)) = \\ &\|f(x^{(i)})\|^2 - 2 f(x^{(i)})^T f(g(x^{(i)})) + \|f(g(x^{(i)}))\|^2 ,
\end{split}
\end{equation}
where the first and third terms are explicit logit regularization terms on $f(x^{(i)})$ and $f(g(x^{(i)}))$, and the logit pairing effect is only determined by the middle inner product. While using an $L^2$ loss is a natural loss for regularization purposes, the pairing term can be improved by considering a more general form:

\begin{equation}
\begin{split}
  &L(f(x^{(i)}; \theta), f(g(x^{(i)}); \theta)) = \\ & h(f(x^{(i)}), f(g(x^{(i)}))) + \beta (\|f(x^{(i)})\|^2 + \|f(g(x^{(i)}))\|^2)
 ,
\end{split}
\end{equation}
where $h$ has the express goal of making the logits more similar (with as little logit regularization as possible), and the regularization terms have been grouped with a controllable weighting factor. There are several natural choices for $h$, such as the Jensen-Shannon divergence, a cosine similarity, or any similarity metric that does not have a significant regularization effect. We have found that simply taking the cross entropy between the distributions induced by the logits was effective -- 
depending on the actual values of the logits, this can either still have a mild squeezing effect (if the logits are very different), a mild expanding effect (if the logits are very similar), or something in between. 

One implementation detail worth noting is that it can be difficult to reason about and set the relative strengths of the pairing loss and adversarial training loss. To that end, we set the strength of the pairing loss $h$ as a constant fraction of the adversarial loss, implemented by setting the coefficient of the loss as a constant multiplied by a non-differentiable version of the ratio between the losses.

By decomposing adversarial logit pairing explicitly into logit pairing and logit regularization terms in this way, adversarial robustness to a 10-step PGD attack improves by an absolute $1.9\%$ over ALP, or $5.6\%$ over PGD-based adversarial training.

\section{Additional experiments}
\label{sec:additional_experiments}
In this section we present additional experiments on three datasets:
The primary two datasets are CIFAR-10 and CIFAR-100, which are datasets for 10-way and 100-way classification, respectively, each with 50,000 examples, on which we evaluate both white-box and black-box adversarial attacks.
Additionally, we evaluate on SVHN~\cite{netzer2011reading}, which has significantly greater scale in training examples (604,388) and whose 10 classes have an uneven distribution, with the most common class roughly $3$ times more common than the least common class.

\subsection{Implementation Details}
\label{sec:implementation_details}
In the experiments throughout this paper on CIFAR-10/100, we used an 18-layer ResNet~\cite{he2016deep}, equivalent to the ``simple'' model of Madry \emph{et al.}~\cite{madry2018towards}, with a weight decay of $2\cdot10^{-4}$ and a momentum optimizer with strength of $0.9$. Standard data augmentation of random crops and horizontal flips was used.
After a warm up period of 5 epochs, the learning rate peaked at 0.1 and decayed by a factor of 10 at 100 and 150 epochs, training for a total of 200 epochs for models not trained on adversarial examples and 101 epochs for models using adversarial training -- adversarial accuracy tends to increase for a brief period of time after a learning rate decay, then quickly drop by a small amount, an empirical finding also echoed by Schmidt \emph{et al.}~\cite{schmidt2018adversarially}. The minibatch size was 128.

Adversarial examples were constrained to a maximum $L^\infty$ norm of $.03$, and all PGD-based attacks used a step size of $0.0078$.
For our implementation of adversarial logit pairing, on CIFAR-10 we used a pairing coefficient of $0.5$ as recommended by \cite{kannan2018adversarial}, and found a larger coefficient of $5.0$ more effective on CIFAR-100.

On SVHN, a smaller 8-layer ResNet was used for computational efficiency due to the scale of the dataset, which is a considerable challenge to overcome with adversarial training. Models were trained for 101 epochs, with a learning rate of .001 for the first 5 epochs, .01 until epoch 80, and .001 afterward. Data augmentation consisted of random crops after an initial 4 pixel padding. Adversarial attacks on SVHN were performed with an $L^\infty$ constraint on the perturbation of $12/255$ with a step size of $3/255$.
All adversarial attacks were constructed using the CleverHans library~\cite{papernot2018cleverhans}, implemented in TensorFlow~\cite{abadi2016tensorflow}, and all experiments were done on two Nvidia Geforce GTX 1080 Ti GPUs.

\begin{table}[t]
  \scriptsize
  \centering
  \caption{White-box accuracy of models on CIFAR-10.}
  \begin{tabular}{c|c|c|c|c|c|}
    \cline{2-6}
       & \multicolumn{5}{|c|}{Adversary}\\ \hline
    \multicolumn{1}{|c|}{Training Method} & Natural & FGSM &
      \begin{tabular}{@{}c@{}}PGD \\ (5 steps)\end{tabular} &
      \begin{tabular}{@{}c@{}}PGD \\ (10 steps)\end{tabular} &
      \begin{tabular}{@{}c@{}}PGD \\ (20 steps)\end{tabular} \\ \hline
    \multicolumn{1}{|c|}{Regular Training} & 87.4\% & 6.1\% & 0.0\% & 0.0\% & 0.0\% \\ \hline
    \multicolumn{1}{|c|}{Label Smoothing} & 86.9\% & 54.7\% & 49.4\% & 41.7\% & 34.4\% \\ \hline
    \multicolumn{1}{|c|}{PGD~\cite{madry2018towards}} & 75.8\% & 50.5\% & 51.0\% & 46.1\% & 45.3\% \\ \hline
    \multicolumn{1}{|c|}{ALP~\cite{kannan2018adversarial}} & 74.0\% & 52.6\% & 53.6\% & 49.1\% & 48.5\% \\ \hline
    \multicolumn{1}{|c|}{\ourmethod\,(ours)} & 68.5\% & 52.8\% & 53.8\% & 51.4\% & 51.0\% \\ \hline
  \end{tabular}
  \label{table:whitebox_cifar10}
\end{table}

\subsection{Towards a more robust model}
Given these forms of logit regularization, perhaps the most natural question is whether they can be combined to create an even more robust model. Thus, in this section we focus exclusively on making a model (and comparable baselines) as robust as possible to PGD-based attacks. In particular, for baseline methods (PGD-based adversarial training~\cite{madry2018towards} and adversarial logit pairing~\cite{kannan2018adversarial}), we opt to train exclusively on adversarial examples, effectively setting $\alpha = 0$ in Equation~\ref{eq:adversarial_training}, which roughly trades off $4-5\%$ accuracy for clean test examples for a similar gain in adversarial performance.

To combine the logit regularization methods together, on CIFAR-10 and CIFAR-100 we use a modest amount of label smoothing ($s = 0.1$) and use VH-mixup~\cite{summers2018improved} on the input examples.
For the logit pairing formulation presented in Section~\ref{sec:decouple}, we found different hyperparameters worked best on our two evaluation datasets. On CIFAR-10, we set $\beta = 10^{-3}$, and set the ratio between the adversarial training loss and the pairing loss to $0.125$, which focuses the loss on keeping adversarial and original examples similar. On CIFAR-100, the more challenging dataset, we use $\beta = 3 \cdot 10^{-5}$ and a ratio of $0.95$, which allows the network to focus more on fitting the data while still maintaining a balance with defending against adversarial examples.
On SVHN, we set $s = 0.2$, use $\beta = 10^{-4}$, employ regular mixup~\cite{zhang2018mixup} (which is more suitable for the imagery of SVHN), and use a ratio of $0.95$, similar to CIFAR-100.
We note that these parameters were not tuned much due to resource constraints. We refer to this combination simply as \ourmethod \, (``\ourmethodfull'').

\vspace{-3mm}

\paragraph{CIFAR-10}
White-box performance on CIFAR-10 is shown in Table~\ref{table:whitebox_cifar10}. \ourmethod\, achieves the highest level of adversarial robustness of the methods considered for all PGD-based attacks, and to the best of our knowledge represents the most robust method on CIFAR-10 to date. However, like other adversarial defenses, this comes at the cost of performance on the original test set, which makes sense -- from the perspective of adversarial training, a clean test image is simply the center of the set of feasible adversarial examples. Nonetheless, it is interesting that the tradeoff between adversarial and non-adversarial performance can continue to be pushed further, with the optimal value of that tradeoff dependent on application, \emph{i.e.} whether worst-case performance is more important than performance on the original unperturbed examples.

\begin{table}[t]
  \scriptsize
  \centering
  \caption{Black-box accuracy of models on CIFAR-10.}
  \begin{tabular}{c|c|c|c|c|c|}
    \cline{2-6}
       & \multicolumn{5}{|c|}{Source}\\ \hline
    \multicolumn{1}{|c|}{Target} & 
      \begin{tabular}{@{}c@{}}Regular \\ Training\end{tabular} &
      \begin{tabular}{@{}c@{}}Label \\ Smoothing\end{tabular} &
        PGD &
        ALP &
        \ourmethod\,(ours) \\ \hline
    \multicolumn{1}{|c|}{Regular Training} & 26.8\% & 32.0\% & 75.1\% & 73.9\% & 67.3\% \\ \hline
    \multicolumn{1}{|c|}{Label Smoothing} & 66.7\% & 67.3\% & 75.6\% & 74.2\% & 67.6\% \\ \hline
    \multicolumn{1}{|c|}{PGD~\cite{madry2018towards}} & 69.8\% & 69.1\% & 58.0\% & 57.9\% & 55.7\% \\ \hline
    \multicolumn{1}{|c|}{ALP~\cite{kannan2018adversarial}} & 69.4\% & 68.8\% & 60.8\% & 59.3\% & 56.4\% \\ \hline
    \multicolumn{1}{|c|}{\ourmethod\,(ours)} & 71.5\% & 70.9\% & 60.5\% & 59.4\% & 54.3\% \\ \hline
  \end{tabular}
  \label{table:blackbox_cifar10}
\end{table}

Next, black-box performance is shown in Table~\ref{table:blackbox_cifar10}. As is standard in most black-box evaluations of adversarial defenses, this is performed by generating adversarial examples with one model (the ``Source'') and evaluating them on a separate independently trained model (the ``Target''). In this experiment, we use a 10-step PGD attack to generate adversarial examples.
As found in other works~\cite{madry2018towards}, the success of a black-box attack depends both on how similar the training procedure was between the source and target models and on the strength of the source model -- for example, \ourmethod\, uniformly results in a stronger black-box attack than ALP~\cite{kannan2018adversarial}, which itself is a uniformly stronger black-box attack than adversarial training with PGD~\cite{madry2018towards}. As such, using \ourmethod\, as the source mildly damages the black-box defenses of PGD and ALP.

Interestingly, label smoothing was fairly effective as a black-box defense, being among the most robust models across all different sources. In particular, label smoothing had the highest minimum performance across sources (over $10\%$ higher than any other method), which is particularly surprising given its near-zero cost compared to the adversarially-trained models.

\vspace{-3mm}
\paragraph{CIFAR-100}

\begin{table}[t]
  \scriptsize
  \centering
  \caption{White-box accuracy of models on CIFAR-100.}
  \begin{tabular}{c|c|c|c|c|c|}
    \cline{2-6}
       & \multicolumn{5}{|c|}{Adversary}\\ \hline
    \multicolumn{1}{|c|}{Training Method} & Natural & FGSM &
      \begin{tabular}{@{}c@{}}PGD \\ (5 steps)\end{tabular} &
      \begin{tabular}{@{}c@{}}PGD \\ (10 steps)\end{tabular} &
      \begin{tabular}{@{}c@{}}PGD \\ (20 steps)\end{tabular} \\ \hline
    \multicolumn{1}{|c|}{Regular Training} & 59.1\% & 2.3\% & 0.1\% & 0.0\% & 0.0\% \\ \hline
    \multicolumn{1}{|c|}{Label Smoothing} & 55.0\% & 11.5\% & 2.3\% & 0.9\% & 0.2\% \\ \hline
    \multicolumn{1}{|c|}{PGD~\cite{madry2018towards}} & 50.2\% & 26.2\% & 27.2\% & 23.9\% & 23.6\% \\ \hline
    \multicolumn{1}{|c|}{ALP~\cite{kannan2018adversarial}} & 44.9\% & 27.3\% & 28.3\% & 26.0\% & 25.9\% \\ \hline
    \multicolumn{1}{|c|}{\ourmethod\,(ours)} & 43.8\% & 28.1\% & 29.1\% & 26.8\% & 26.6\% \\ \hline
  \end{tabular}
  \label{table:whitebox_cifar100}
\end{table}

White-box performance on CIFAR-100 is presented in Table~\ref{table:whitebox_cifar100}.
Again, we find that \ourmethod\, achieves the highest level of adversarial robustness to all adversarial attacks, with ALP also strictly better than PGD-based adversarial training. Interesting, though, we find that label smoothing completely fails to all attacks on CIFAR-100, behaving almost completely differently than on CIFAR-10. Although examining this was not the goal of our work, this does highlight the importance of evaluating proposed adversarial defenses on multiple datasets.

The corresponding results for black-box attacks on CIFAR-100 are shown in Table~\ref{table:blackbox_cifar100}, where we again find clear differences across datasets. This time, the difference is that all methods perform much more similarly to one another, with the exception of clear differences of transferring from PGD-trained models to non-adversarially trained models.

\paragraph{SVHN}
We demonstrate performance against white-box attacks on SVHN in Table~\ref{table:whitebox_svhn}.
Despite the large differences in dataset scale, image type, and imbalance in class frequencies, we again find that our method, \ourmethod, is the most robust of all appoaches on every adversarial attack, with patterns in the performance of each defense very similar to the other datasets, providing additional evidence that our methods and insights are generalizable.

\subsection{Evaluating Stronger Attacks}
\label{sec:stronger_attacks}
Evaluating adversarial defenses is difficult to do correctly -- since evaluating against any attack merely provides an upper bound on adversarial robustness, it is critical to evaluate on the strongest attacks available to make the bound as tight as possible.
Furthermore, care must be taken to avoid gradient masking or obfuscated gradients~\cite{athalye2018obfuscated}, which can lead to a false sense of security.

Here we evaluate white-box performance on CIFAR-10 with two very strong attacks: a 1,000-step PGD adversary (the same attack that ALP succumbed to on ImageNet), and SPSA~\cite{uesato2018adversarial}, a gradient-free attack that is effective at uncovering gradient masking. Results are given in Table~\ref{table:whitebox_strong_cifar10}. Note that SPSA is evaluated against a representative 1,000-image sample of the evaluation set for efficiency, since a full evaluation would take roughly 90 hours, and that we use the same evaluation settings as provided in ~\cite{uesato2018adversarial}.

We find nearly no difference when going from a 20-step to a 1,000-step PGD attack for all methods except for label smoothing, which loses most of its robustness.  This suggests that label smoothing, while providing only a mild amount of worst-case adversarial robustness, can actually make the adversarial optimization problem much more challenging, which we believe is also the underlying reason for its effectiveness against black-box attacks.
Based on this conjecture, we also evaluated label smoothing as a black-box defense with a 1,000-step PGD attack, where we have found a much smaller drop in performance, going from $67.3\%$ to $60.8\%$, confirming that label smoothing still has its place in black-box defenses.  The exact mechanism by which label smoothing makes the search for adversarial examples more difficult, however, remains elusive, which we think is an interesting avenue for further research.

On the other hand, an SPSA attack removes some of the difference in robustness between PGD, ALP, and \ourmethod\,.  While this illustrates that ALP and \ourmethod\, are likely doing some type of gradient masking in a way that PGD cannot detect, even with 1,000 iterations, it also illustrates that there is a real gain in adversarial robustness even when considering strong gradient-free attacks.

\begin{table}[t]
  \scriptsize
  \centering
  \caption{Black-box accuracy of models on CIFAR-100.}
  \begin{tabular}{c|c|c|c|c|c|}
    \cline{2-6}
       & \multicolumn{5}{|c|}{Source}\\ \hline
    \multicolumn{1}{|c|}{Target} & 
      \begin{tabular}{@{}c@{}}Regular \\ Training\end{tabular} &
      \begin{tabular}{@{}c@{}}Label \\ Smoothing\end{tabular} &
        PGD &
        ALP &
        \ourmethod\,(ours) \\ \hline
    \multicolumn{1}{|c|}{Regular Training} & 33.3\% & 31.7\% & 50.1\% & 44.7\% & 43.8\% \\ \hline
    \multicolumn{1}{|c|}{Label Smoothing} & 37.0\% & 31.6\% & 50.0\% & 44.8\% & 43.9\% \\ \hline
    \multicolumn{1}{|c|}{PGD~\cite{madry2018towards}} & 37.6\% & 34.0\% & 33.4\% & 32.5\% & 32.4\% \\ \hline
    \multicolumn{1}{|c|}{ALP~\cite{kannan2018adversarial}} & 39.7\% & 36.6\% & 33.6\% & 31.5\% & 32.1\% \\ \hline
    \multicolumn{1}{|c|}{\ourmethod\,(ours)} & 37.7\% & 33.4\% & 34.5\% & 33.0\% & 31.3\% \\ \hline
  \end{tabular}
  \label{table:blackbox_cifar100}
\end{table}

\begin{table}[t]
  \scriptsize
  \centering
  \caption{White-box accuracy of models on SVHN.}
  \begin{tabular}{c|c|c|c|c|c|}
    \cline{2-6}
       & \multicolumn{5}{|c|}{Adversary}\\ \hline
    \multicolumn{1}{|c|}{Training Method} & Natural & FGSM &
      \begin{tabular}{@{}c@{}}PGD \\ (5 steps)\end{tabular} &
      \begin{tabular}{@{}c@{}}PGD \\ (10 steps)\end{tabular} &
      \begin{tabular}{@{}c@{}}PGD \\ (20 steps)\end{tabular} \\ \hline
    \multicolumn{1}{|c|}{Regular Training} & 96.7\% & 14.8\% & 0.0\% & 0.0\% & 0.0\% \\ \hline
    \multicolumn{1}{|c|}{Label Smoothing} & 97.0\% & 50.8\% & 15.1\% & 4.4\% & 1.3\% \\ \hline
    \multicolumn{1}{|c|}{PGD~\cite{madry2018towards}} & 85.3\% & 50.5\% & 47.3\% & 39.6\% & 38.0\% \\ \hline
    \multicolumn{1}{|c|}{ALP~\cite{kannan2018adversarial}} & 82.5\% & 49.6\% & 47.1\% & 40.0\% & 38.6\% \\ \hline
    \multicolumn{1}{|c|}{\ourmethod\,(ours)} & 83.6\% & 51.5\% & 48.4\% & 40.9\% & 39.4\% \\ \hline
  \end{tabular}
  \label{table:whitebox_svhn}
\end{table}

\begin{table}[t]
  \scriptsize
  \centering
  \caption{Evaluating models against the strongest white-box attacks on CIFAR-10. SPSA is evaluated on a 1,000-image ($10\%$) subsample, and a 20-step PGD attack is provided for context.}
  \begin{tabular}{c|c|c|c|}
    \cline{2-4}
       & \multicolumn{3}{|c|}{Adversary}\\ \hline
    \multicolumn{1}{|c|}{Training Method} &
      \begin{tabular}{@{}c@{}}PGD \\ (20 steps)\end{tabular} &
      \begin{tabular}{@{}c@{}}PGD \\ (1,000 steps)\end{tabular} &
			SPSA~\cite{uesato2018adversarial} \\ \hline
    \multicolumn{1}{|c|}{Regular Training} & 0.0\% & 0.0\% & 0.0\% \\ \hline
    \multicolumn{1}{|c|}{Label Smoothing} & 34.4\% & 7.2\% & 8.9\% \\ \hline
    \multicolumn{1}{|c|}{PGD~\cite{madry2018towards}} & 45.3\% & 45.2\% & 45.4\% \\ \hline
    \multicolumn{1}{|c|}{ALP~\cite{kannan2018adversarial}} & 48.5\% & 48.3\% & 46.1\% \\ \hline
    \multicolumn{1}{|c|}{\ourmethod\,(ours)} & 51.0\% & 51.1\% & 47.9\% \\ \hline
  \end{tabular}
  \label{table:whitebox_strong_cifar10}
  \vspace{-4mm}
\end{table}

\section{Discussion}
In this work, we have shown the usefulness of logit regularization for improving the robustness of neural models of computer vision to adversarial examples. We first presented an analysis of adversarial logit pairing, showing that roughly half of its improvement over adversarial training can be attributed to a non-obvious logit regularization effect. Based on this, we investigated two other forms of logit regularization, demonstrating the benefits of both, and then presented an alternative method for adversarial logit pairing that more cleanly decouples the logit pairing and logit regularization effects while also improving performance.

By combining these logit regularization techniques together, we were able to create both a stronger defense against white-box PGD-based attacks and also a stronger attack against PGD-based defenses, both of which come at almost no additional cost to PGD-based adversarial training.
We also demonstrated the surprising strength of label smoothing as a black-box defense and its paradoxical weakness to highly-optimized white-box attacks.

We anticipate that future work will push the limits of logit regularization even further to improve defenses against adversarial examples, possibly drawing on techniques originally devised for other purposes~\cite{pereyra2017regularizing}.
We also hope that these investigations will yield insights into training adversarially-robust models without the overhead of multi-step adversarial training, an obstacle that has made it challenge to scale up adversarial defenses to larger datasets without very large computational budgets.

{\small
\bibliographystyle{ieee}
\bibliography{references}
}

\end{document}